\PassOptionsToPackage{full}{textcomp}
\documentclass[conference]{IEEEtran}
\IEEEoverridecommandlockouts
\usepackage[normalem]{ulem} 
\usepackage[boxed,ruled,vlined,linesnumbered]{algorithm2e}

\usepackage{pifont}
\let\oldding\ding 
\renewcommand{\ding}[2][1]{\scalebox{#1}{\oldding{#2}}} 
\usepackage{enumitem}

\usepackage{cite,xspace,comment}
\usepackage{amsmath,amsthm,amssymb,amsfonts}
\usepackage[ruled,vlined]{algorithm2e}
\usepackage{graphicx,subfigure}
\usepackage{textcomp}
\usepackage[dvipsnames]{xcolor}
\usepackage{multirow}
\usepackage{cleveref}
\crefname{figure}{Fig.}{Figs.}
\Crefname{figure}{Fig.}{Figs.}
\usepackage{siunitx}
\usepackage{xurl}
\usepackage{booktabs}
\usepackage{newtxtext}

\usepackage[acronym]{glossaries}

\newacronym{ml}{ML}{Machine Learning}
\newacronym{fl}{FL}{Federated Learning}
\newacronym{he}{HE}{Homomorphic Encryption}
\newacronym{dlg}{DLG}{Deep Leakage from Gradients}
\newacronym{fedavg}{FedAvg}{Federated Averaging}
\newacronym{fedavghe}{\textsf{FedAvg-HE}}{FedAvg embedded with HE}
\newacronym{fedfa}{FedFA}{Federated Learning with Feature Anchors}
\newacronym{fedfahe}{\textsf{FedFA-HE}}{FedFA embedded with HE}
\newacronym{fedfashe}{FedFA-Selective-HE}{FedFA embedded with Selective HE}
\newacronym{nameNovel}{\textsf{Alt-FL}}{Alternating Federated Learning}
\newacronym{fe}{\emph{FE}, $\rho{=}0$}{\emph{FE}, $\rho{=}0$}
\newacronym{s-he}{\emph{S-HE}, $\rho{=}0$}{\emph{FE}, $\rho{=}0$}
\newacronym{has-he}{\emph{Ext, S-HE}, $\rho{=}0$}{\emph{Extended, S-HE}, $\rho{=}0$}

\newcommand{\BB}{}
\newcommand{\BBB}{}

\newcommand{\etal}{\emph{et al.}\xspace}
\newcommand{\eg}{\emph{e.g.,}\xspace}
\newcommand{\ie}{\emph{i.e.,}\xspace}

\def\BibTeX{{\rm B\kern-.05em{\sc i\kern-.025em b}\kern-.08em
    T\kern-.1667em\lower.7ex\hbox{E}\kern-.125emX}}

\begin{document}

\title{Integrating Homomorphic Encryption and Synthetic Data in FL for Privacy and Learning Quality
}

\author{\IEEEauthorblockN{Yenan Wang}
\IEEEauthorblockA{Computer Science and Engineering \\
Chalmers University of Technology\\
Gothenburg, Sweden 
}
\and
\IEEEauthorblockN{Carla Fabiana Chiasserini}
\IEEEauthorblockA{Computer Science and Engineering \\
Chalmers University of Technology\\
Gothenburg, Sweden
}
\and
\IEEEauthorblockN{Elad Michael Schiller}
\IEEEauthorblockA{Computer Science and Engineering \\
Chalmers University of Technology \\
Gothenburg, Sweden} 
}

\maketitle

\begin{abstract}
Federated learning (FL) enables collaborative training of machine learning models without sharing sensitive client data, making it a cornerstone for privacy-critical applications. However, FL  faces the dual challenge of ensuring learning quality and robust privacy protection while keeping resource consumption low, particularly when using computationally expensive techniques such as homomorphic encryption (HE).  
In this work, we enhance an FL process that preserves privacy using HE by integrating it with synthetic data generation and an interleaving strategy. Specifically, our solution, named \gls{nameNovel},  consists of alternating between local training with authentic data (authentic rounds) and local training with synthetic data (synthetic rounds) and transferring the encrypted and plaintext model parameters on authentic and synthetic rounds (resp.).   
Our approach improves learning quality (\eg model accuracy) through datasets enhanced with synthetic data, preserves client data privacy via \acrshort{he}, and keeps manageable encryption and decryption costs through our interleaving strategy.  
We evaluate our solution against data leakage attacks, such as the DLG attack, demonstrating robust privacy protection. Also,   \gls{nameNovel} provides 13.4\%  higher model accuracy and decreases HE-related costs by up to 48\% with respect to Selective \acrshort{he}. 
\end{abstract}


\section{Introduction}
\label{sec:introduction}

\gls{fl}~\cite{DBLP:conf/aistats/McMahanMRHA17,DBLP:journals/tist/YangLCT19} has gained significant attention as it allows multiple clients to train a global model locally and share only the model updates with a central server, which aggregates them to update the global model.
\acrshort{fl} has thus become indispensable for a wide range of applications where it is crucial to keep the clients' local data private, \eg healthcare, 
banking, 
and smart cities. 
However, besides safeguarding privacy-sensitive information, these applications must manage vast amounts of data distributed across multiple sources while addressing storage and communication constraints.

It follows that, despite recent advances, \acrshort{fl} still poses several challenges. 
Notably, even if \acrshort{fl} shares and aggregates local models, avoiding raw data transmission, transferring model parameters multiple times during the training iterations yields significant overhead and introduces vulnerabilities. 
The latter issue has been highlighted, \eg in the seminal work by Zhu \etal~\cite{DLG}, which presents an attack, called \gls{dlg}, performing an inversion attack on the shared parameter gradients between the server and clients to recover the clients’ training data. 

\noindent\textbf{Related Work.} 
As model updates may leak privacy-sensitive client data~\cite{DBLP:series/lncs/12500}, several privacy-enhancing solutions have been proposed since \gls{fedavg}~\cite{DBLP:conf/aistats/McMahanMRHA17} was introduced. 
For example, Bonawitz \etal~\cite{DBLP:journals/corr/BonawitzIKMMPRS16} developed a secure aggregation protocol using double-masking to protect against honest-but-curious servers, while Konečn\`y \etal~\cite{DBLP:journals/corr/KonecnyMYRSB16} proposed structured and sketched updates to reduce communication costs.
In this study, we adopt FedAvg for model updating.

\gls{he}~\cite{DBLP:conf/asiacrypt/CheonKKS17} is a cryptographic method that enables computation on encrypted data without decryption, offering strong privacy guarantees.
Despite its benefits in \acrshort{fl}, \acrshort{he} can incur substantial computational and communication overhead due to ciphertext expansion and costly operations.
To address this, Cheon \etal~\cite{DBLP:conf/asiacrypt/CheonKKS17} introduced a scheme with reduced ciphertext size, which Jin \etal~\cite{DBLP:journals/corr/abs-2303-10837} leveraged in FedML-HE by selectively encrypting sensitive parameters.
Their approach also supports tunable encryption ratios, enabling adaptation to different privacy requirements.

Differential Privacy (DP)~\cite{DLwithDP} is another widely used technique that offers provable privacy guarantees through a rigorous theoretical framework. In machine learning, DP typically adds calibrated noise to gradients, which can impair model performance. For example, Wei \etal~\cite{FLwithDP} introduced the NbAFL framework, combining client-level DP with FL, and showed that stronger privacy degrades convergence. In contrast, our study prioritizes the trade-off between computational overhead and model accuracy, and therefore adopts Homomorphic Encryption (HE) exclusively for privacy protection.

Beyond DLG, stronger attacks have emerged~\cite{DBLP:conf/nips/GeipingBD020,DBLP:conf/cvpr/YinMVAKM21,DBLP:conf/iclr/FowlGCGG22,DBLP:conf/sp/ZhaoSEEAB24}.
But, these attacks either need extra information like batch normalization statistics~\cite{DBLP:conf/nips/GeipingBD020,DBLP:conf/cvpr/YinMVAKM21} (to maximize attack effectiveness) or assume a malicious server~\cite{DBLP:conf/iclr/FowlGCGG22,DBLP:conf/sp/ZhaoSEEAB24}.
Therefore, we focus on the DLG attack, which only requires gradient information.

As far as learning quality is concerned, several studies have analyzed the convergence of an FL process and the training or testing loss that can be achieved depending on the characteristics and quality of the data that clients use for their local training (see, \eg~\cite{scaffold,iclear24_poster_bounds}). Another body of work has instead focused on generating synthetic data and its benefits for training machine learning (ML) models  \cite{Singh_2021,lu2024syntheticdata}. Notably, enhancing real-world data with synthetic data can greatly benefit learning quality, as more balanced datasets can be obtained. However, when synthetic data is used, more data needs to be processed, and the global model may take longer to converge, which may result in exceedingly increased training time. Consequently, the FL process may require more bandwidth and computational resources.

\noindent\textbf{Open Issues and Proposed Solution.}   
We identify the following two main challenges. 
{\em First}, protecting privacy-sensitive data in \acrshort{fl} systems via \acrshort{he} introduces computational overhead due to encryption and decryption at each client at every training round. 
{\em Second}, synthetic data can be integrated into the FL training process to improve the level of learning quality achieved but at the expense of an increased training cost. As noted above, such a cost increase is further amplified by  \acrshort{he} whenever the latter is applied to training rounds plainly. 

We tackle the above challenges as follows:

\noindent $\bullet$
\emph{Ensuring privacy protection for FedAvg.} 
Our method leverages Selective \acrshort{he} to secure the privacy of clients' data during the communication process with the \acrshort{fl} server; 

\noindent $\bullet$ 
\emph{Incorporating synthetic data.} We incorporate synthetic datasets into the \acrshort{fl} process for local training at the clients to obtain more balanced client datasets. The aim is to improve learning quality and the efficiency of Selective \acrshort{he} by balancing the clients' datasets.

\noindent $\bullet$ 
\emph{Interleaved use of authentic and synthetic data.}  
    To reduce the bandwidth and computational costs of \acrshort{he}, we integrate the use of synthetic data in the training process by alternating \emph{authentic} and \emph{synthetic} rounds. 
    During authentic rounds, the model is trained locally at the clients using the actual client's data, and the locally trained model is encrypted before being transmitted. During synthetic rounds, instead, the model is trained at the clients using synthetic data and is then sent unencrypted. 
    We underline that the model obtained from an authentic round is retrained on synthetic data in the next round.  

\noindent $\bullet$ 
\emph{Tunable ratio between authentic and synthetic rounds.} 
    Increasing the proportion of synthetic rounds reduces the per-round resource consumption due to encryption and decryption while improving learning quality. However, the model convergence time and accuracy performance need to be monitored, as underlined in \cite{debbah2024}.  
    To overcome this issue, our approach introduces a tunable parameter, $\rho$, that controls the ratio between authentic and synthetic rounds, allowing us to trade off system constraints with performance requirements dynamically.

\begin{figure}[htbp]
    \centering
    \BB
    \includegraphics[width=0.9\linewidth]{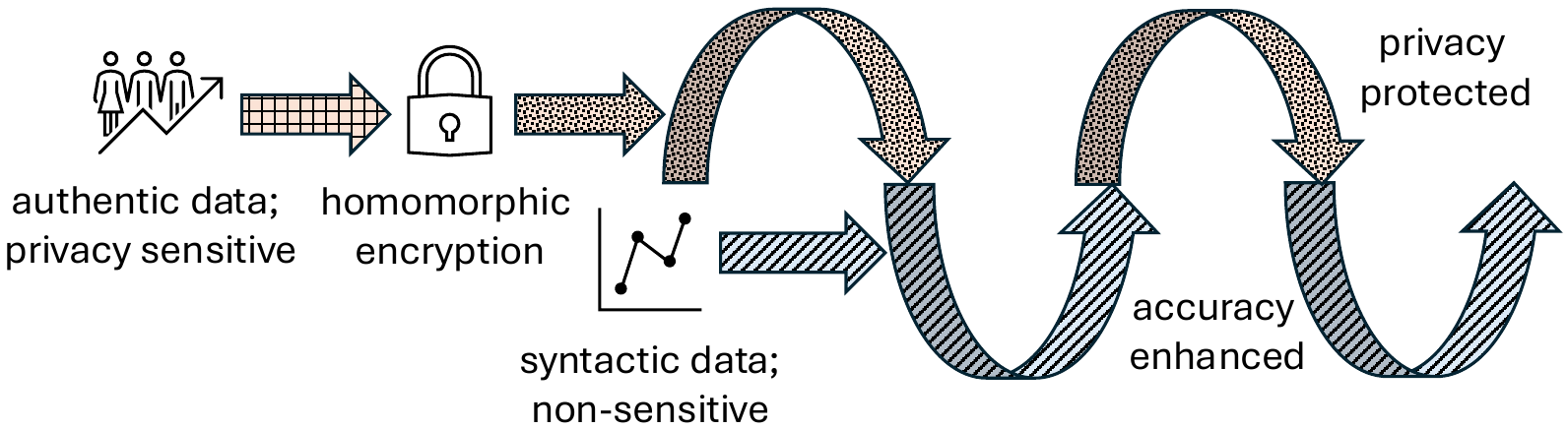}
    \caption{An overview of the proposed \acrfull{nameNovel}.}
    \label{fig:conceptFigure}
    \BB
\end{figure}

\noindent\textbf{Our Contribution.}  
We enhance \acrshort{fl} by introducing \gls{nameNovel}, which protects privacy and improves learning quality using \acrshort{he} and synthetic data while keeping the resource consumption of encryption and decryption low. 
\acrshort{nameNovel}, depicted in Fig.\,\ref{fig:conceptFigure} is the first framework that addresses the limitations of Selective \acrshort{he} with interleaved authentic and synthetic data rounds in \acrshort{fl}.  

\noindent\textbf{Validation and Impact.}  
Our results highlight that:  

\noindent $\bullet$ \acrshort{nameNovel} can increase model accuracy by 13.4\%;  

\noindent $\bullet$ \acrshort{nameNovel}  counteracts  \acrshort{dlg} attacks, achieving privacy performance comparable to or higher than state-of-the-art methods; 
 
\noindent $\bullet$  \acrshort{nameNovel}   reduces encryption and decryption overheads by up to 48\% with respect to the non-interleaving case.  

These findings show that our approach increases learning quality and maintains robust privacy protection while significantly limiting the bandwidth and computing resource consumption associated with \acrshort{he} encryption and decryption.  
By addressing critical challenges in \acrshort{fl}, \acrshort{nameNovel} facilitates more scalable deployment of \acrshort{fl} in privacy-critical applications.

To allow for the reproducibility of our results and foster further development, the implementation of our solution is publicly available on GitHub~\cite{ourRepo}.

\section{Preliminaries}\label{sec:methods}

Our work integrates FedAvg with HE using an interleaved strategy (described in~\Cref{sec:ourApproach}).

\noindent \textbf{FL and FedAvg.}
Our FL setup consists of $N$ clients, each with its own private dataset $d_i {\in} \mathcal{D}$. 
These clients collaborate with a central server to train a shared global model $w^t$ over multiple communication rounds $t$.
In  FedAvg~\cite{DBLP:conf/aistats/McMahanMRHA17}, each client trains a local model $w_i^t$ using its own data.
The server then aggregates such models into a global model $w^t$ by computing a weighted average $w^t {=} \sum_{i=1}\frac{|d_i|}{|\mathcal{D}|}w_i^t$, where the weight for each client's model is proportional to the size of its dataset relative to the total data across all clients.

\noindent \textbf{Embedding Homomorphic Encryption.}
To enhance privacy within FedAvg,  Selective \acrshort{he}~\cite{DBLP:journals/corr/abs-2303-10837} can be used.
In this approach, only sensitive data is encrypted, reducing computational overhead while preserving privacy. 
Specifically, selective \acrshort{he}~\cite{DBLP:journals/corr/abs-2303-10837} balances encryption overhead and privacy by encrypting only high-risk model parameters based on a sensitivity metric. This approach ensures efficient yet secure parameter sharing in \acrshort{fl}, especially when the heterogeneity of the clients is low. 
	
The encryption mask \( M \) is a binary indicator identifying which model parameters should be encrypted. 
Once the server aggregates sensitivity values from all clients, it applies a predefined threshold \( T \) when computing the mask elements \( M(w_m) \). 
Only parameters with \( M(w_m) {=} 1 \) are encrypted, reducing computational overhead while preserving privacy. 
We define the encryption ratio as $\eta {=} \tfrac{\sum_m M_(w_m)}{|M|}$, where 
$|M|$ is the number of elements in $M$. 
For simplicity, our implementation calculates the mask from the encryption ratio directly.

\noindent \textbf{Attacker Model and \gls{dlg} Attack.} 
In line with Hardy \etal~\cite{DBLP:journals/corr/abs-1711-10677}, we adopt the \emph{honest-but-curious model}, which assumes that: (i) clients faithfully execute the \acrshort{fl} algorithm as intended; (ii) no collusion occurs between clients; and (iii) clients may still seek to infer as much information as possible from what other clients reveal. 
While each client's dataset remains confidential, the set of features and their types are assumed to be public.

The \acrshort{dlg} attack, introduced by Zhu \etal~\cite{DLG}, targets \acrshort{fl} systems by exploiting gradient sharing. 
The goal is to infer privacy-sensitive information from gradients exchanged between clients and the server, exposing pixel-level details of privacy-sensitive training images. 
\acrshort{dlg} operates by initializing the parameters of a dummy model at the attacker with random values and iteratively minimizing the difference between the dummy gradients and the authentic gradients computed during the training of the dummy model. 
\acrshort{dlg} is relatively easy to implement since it only requires having shared gradients without additional knowledge of the dataset. 

Note that \acrshort{fedavg} clients do not explicitly share their gradients with the server. 
Thus, we assume the attacker can infer gradients only from the plain-text model parameters in a communication round during \acrshort{fedavg}'s training process. 
Therefore, our implementation of the \acrshort{dlg} attack assumes that the attacker can compute the gradients from the most recent plain-text  model parameters sent by each client on the training round at which the attack is launched. 
In practice,  the attacker can observe gradients for unencrypted parameters, but cannot do so for encrypted parameters, as they appear as random numbers.
Additionally, we assume that the gradients are calculated for a single image instead of a batch of multiple images.
This is done to force recovery of the training image in a reasonable amount of time since, as shown in \cite{DLG}, such a time quickly becomes exceedingly high as the batch size~grows.

\section{\acrfull{nameNovel}}\label{sec:ourApproach}

Using privacy-preserving techniques like \acrshort{he} in \acrshort{fl}  introduces significant overhead due to encryption and decryption operations, leading to performance bottlenecks. 
To address this issue, we envision an interleaving approach that integrates synthetic data into the \acrshort{fl} process. The aim of enhancing the clients' local datasets with synthetic data is twofold: {\em (i)} to improve the learning quality, \ie the model accuracy, and {\em (ii)} to reduce the \acrshort{he}-related overhead, as better detailed below.

\noindent \textbf{Key Concepts.}  
Our proposed approach utilizes synthetic client datasets that are statistically similar but disjointed from authentic datasets. 
The training process alternates between authentic rounds, in which clients use authentic (real-world) data, and synthetic rounds, in which clients use synthetic data.   
In authentic rounds, client models trained on authentic data are encrypted using Selective \acrshort{he} and transmitted to the server to ensure privacy. 
In contrast, during synthetic rounds, client models trained on synthetic data are transmitted unencrypted, thus reducing bandwidth and computational costs.  
At every round, a client trains the model received from the aggregator at the previous round; for example, a synthetic round trains the model that the aggregator has just output by combining those the clients trained at the previous (authentic) round. 

Importantly, using synthetic data allows for an improved accuracy level \cite{Singh_2021,lu2024syntheticdata}. Furthermore, the above method reduces the \acrshort{he} overhead by {\em (i)} implementing interleaving rounds and {\em (ii)} applying, during authentic rounds, Selective \acrshort{he} on models that  (overall) have been trained with more balanced data thanks to the synthetic data used by each client. 

A key feature of our approach is the tunable interleaving ratio, $ \rho {\in} [0,1] $, which defines the ratio of synthetic to total number of rounds, \ie $\rho{=}\rho_{syn}/ \rho_{tot}$.
This ratio can be adjusted to balance system performance with privacy requirements.

\noindent \textbf{Algorithm Description.} 
\Cref{alg:fl-he-inter} integrates Selective \acrshort{he} within the \acrshort{fl} process using an interleaving approach to balance privacy protection and computational efficiency. The input parameters in \Cref{ln:input} include the encryption mask \(m\), which identifies sensitive parameters for encryption, \(d_{i,a}\),  representing the authentic dataset for client \(i\), and \(d_{i,s}\) representing the synthetic dataset for client \(i\).

In~\Cref{ln:forRound}, the algorithm starts each round by determining if it is authentic based on the interleaving ratio, $\rho$, calculated as \(t \bmod \rho_{tot} {<} \rho_{tot} {-} \rho_{syn}\)
(\Cref{ln:isAuth}). This condition dictates whether clients use authentic or synthetic datasets and whether models are encrypted or transmitted in plaintext.

Clients first check if the incoming model \(w^{t-1}\) is encrypted (\Cref{ln:testEnc}), decrypt it if necessary (\Cref{ln:ifEnc,ln:elseDec}), and assign the appropriate dataset (\Cref{ln:ifAuthDataset,ln:elseDataset}).
They then train their models locally (\Cref{ln:trainModel}). 
In authentic rounds, models are encrypted and sent to the server (\Cref{ln:encryptModel,ln:sendEncModel}), while in synthetic rounds, unencrypted models are transmitted (\Cref{ln:sendPlainModel}). 
The server aggregates received models (\Cref{ln:aggregateModels}) and applies the mask if the round is authentic (\Cref{ln:aggUsingMask}). The aggregated model is then broadcast to all clients (\Cref{ln:broadcastModel}).

\noindent \textbf{Anticipated Benefits.} 
We outline the expected performance and advantages of the proposed solution.  

\noindent $\bullet$ \emph{Improved learning accuracy.} Drawing on, \eg  \cite{Singh_2021,lu2024syntheticdata}, we expect \acrfull{nameNovel} to increase the accuracy level achieved through the FL process, thanks to the use of synthetic data for the local training performed by the clients on synthetic rounds. 
	
\noindent $\bullet$  \emph{Privacy preservation.}
As explained in~\Cref{sec:methods}, the \acrshort{dlg} attacker can infer the model gradients at the end of each synthetic (\ie plain-text) communication round. 
However, Selective \acrshort{he} protects the privacy of the client models on authentic rounds so that the latter cannot lead to the exposure of privacy-sensitive information.  
During synthetic rounds (\Cref{ln:elseDataset}), although models are transferred in plain-text, privacy leakage is still avoided since only synthetic data is used and every item of the synthetic datasets (\Cref{ln:elseDataset}) is disjoint from any item in the authentic datasets (\Cref{ln:ifAuthDataset}). 

\noindent $\bullet$  \emph{Cost reduction.}
We expect \acrshort{nameNovel} to keep the bandwidth consumption and computational overhead of \acrshort{he} low by applying Selective \acrshort{he} only during authentic rounds and on models that (overall) have been trained using more balanced datasets. Indeed, local datasets are enhanced with synthetic data that can help mitigate the heterogeneity across clients' datasets. 
Specifically, let $n$ be the number of rounds necessary for the convergence of the FedAvg process. Also, let $\text{cost}_{\textsf{FedAvg-HE}}(n)$ be the total bandwidth and computational cost of FedAvg embedded with Selective \acrshort{he}, when $n$ training rounds are performed. For each round, this cost includes two components: $\text{cost}_{\textsf{FedAvg}}$, the cost of FedAvg, and $ \text{cost}_{\textsf{HE}}$, the cost of \acrshort{he}: \BB
\begin{equation}
\BB
\text{cost}_{\textsf{FedAvg-HE}}(n) = n\cdot (\text{cost}_{\textsf{FedAvg}} + \text{cost}_{\textsf{HE}})\,.
\label{eq:simple}
\end{equation}

Assuming equal training data size (in bytes) for authentic and synthetic rounds, the total cost of \acrshort{nameNovel} with interleaving ratio~$\rho$ is:
\begin{equation}
\text{cost}_{\acrshort{nameNovel}}(n) = \hat{n}(\rho) \cdot \text{cost}_{\textsf{FedAvg}} + (1 - \rho) \cdot \hat{n}(\rho) \cdot \widehat{\text{cost}}_{\textsf{HE}}
\label{eq:next}
\end{equation}
Here, $\hat{n}(\rho)$ is the number of rounds to convergence, expected to exceed $n$, and $\widehat{\text{cost}}_{\textsf{HE}}$ is the per-round \acrshort{he} cost in \acrshort{nameNovel}, which may differ from $\text{cost}_{\textsf{HE}}$ due to different model structures and encryption mask selections compared to \acrshort{fedavg}. 
Given these dependencies, the actual cost of \acrshort{nameNovel} must be empirically evaluated (see \Cref{sec:results}).

\begin{algorithm}[t!]
\begin{smaller}
	\caption{\label{alg:fl-he-inter}\smaller{The proposed \acrshort{nameNovel} algorithm}}
\textbf{input:} {encryption mask: $m$, authentic datasets for client $i$: $d_{i,a}$, and synthetic datasets for client $i$: $d_{i,s}$}\label{ln:input}\;
	
	\For{\textup{round} $t = 1,\dots,R$}{ \label{ln:forRound}
		\texttt{// $isAuth$ holds if round $t$ is authentic}\;
		$\text{isAuth} \gets t \bmod \rho_{syn} < \rho_{tot}-\rho_{syn}$ \textbf{where} $\rho{=}\frac{\rho_{syn}}{\rho_{tot}}$ is the interleaving ratio\label{ln:isAuth}\; 
		
		\texttt{// client-side code}\;
		\textbf{upon} model $w^{t-1}$ \textbf{arrival to} client $i \in S$ \Begin{ \label{ln:clientBegin}
			\texttt{// decrypt the arriving model, if needed}\;
			$\text{isEnc} \gets$ test if $w^{t-1}$ is encrypted\label{ln:testEnc}\; 
			\lIf{$isEnc$}{$w^t_i \gets Dec(w^{t-1},\,m)$\label{ln:ifEnc}} 
			\lElse{$w^t_i \gets w^{t-1}$} \label{ln:elseDec}
			
			\texttt{// assign the appropriate dataset for this round}\;
			\lIf{$isAuth$}{$d_i \gets$ $d_{i,a}$\label{ln:ifAuthDataset}} 
			\lElse{$d_i \gets d_{i,s}$\label{ln:elseDataset}} 
			
			\texttt{// client model training}\;
			$w^t_i \gets$ train local model $w^t_i$ with dataset $d_i$\label{ln:trainModel}\; 
			
			\texttt{// client model transmission}\;
			\If{\textup{isAuth \textbf{is} true}}{ \label{ln:ifAuthTransmission}
				$Enc(w^t_i,\,m) \gets$ encrypt $w^t_i$\label{ln:encryptModel}\; 
				\textbf{send} $Enc(w^t_i,\,m)$ to server\label{ln:sendEncModel}\; 
			}\lElse{\textbf{send} $w^t_i$ to server\label{ln:sendPlainModel}} 
		} \label{ln:clientEnd}
		
		\texttt{// server-side code}\;
		\textbf{upon arrival of} model $w^t_i$ \textbf{for any} $i \in S$ \Begin{ \label{ln:serverBegin}
			\texttt{// aggregate arriving models}\;
			$W^t \gets$ the set of all received client models\; \label{ln:aggregateModels}
			\If{\textup{isAuth \textbf{is} true}}{ \label{ln:ifAuthAggregate}
				$w^t \gets$ aggregate $W^t$ using mask~$m$\label{ln:aggUsingMask}\;
			}\lElse{$w^t \gets$ aggregate $W^t$} \label{ln:aggWithoutMask}
			
			\texttt{// broadcast the aggregated model}\;
			\lForEach{client $i \in S$}{\textbf{send} $w^t$ to client $i$} \label{ln:broadcastModel}
		} \label{ln:serverEnd}
	} \label{ln:endFor}
\end{smaller}	
\BB
\end{algorithm}

\section{Experiment Settings}\label{sec:evaluation}

We conduct our experiments by varying the encryption ratio $\eta$ (with 0 signifying no encryption and 1 for full model encryption) and interleaving ratio, $\rho$ (specifically, we focus on 0, 0.25, and 0.5, \ie 0\%, 25\%, and 50\%, resp., of all rounds are synthetic).
Note that the case with an interleaving ratio of 0 is also denoted as the non-interleaving case.

\noindent\textbf{Model and Training.}
We focus on a classification task, leveraging  the  LeNet-5 network architecture~\cite{DBLP:journals/pieee/LeCunBBH98}.  The model is trained using  the CIFAR-10 dataset~\cite{CIFAR} and  three  clients, as in~\cite{DBLP:journals/corr/abs-2303-10837}. 
We simplify the synthetic data generation process since we focus on the novelty of our proposal, and there are techniques for providing such synthetic data, \eg~\cite{dockhorn2023differentially,DPGAN,PPGAN}.
Before distributing the data across the clients, we split the training dataset in half, where one half is regarded as \emph{authentic data} and the other is considered as \emph{synthetic data}. 
The authentic data is distributed to each client based on a Dirichlet distribution with the parameter $\alpha{=}0.5$ to simulate non-IID data, whereas the synthetic data are IID.
Each training process uses an Intel Xeon Platinum 8358 CPU, with 4 cores allocated. 
We train the model 10 times for each configuration of encryption and interleaving ratios until convergence, averaging the results after excluding the best and worst performing runs to remove outliers.
In our experiments, we apply a moving average (window size 12) to smooth the accuracies, and we declare convergence when the smoothed accuracies fail to increase by more than $0.1\%$ for 10 consecutive rounds. 
We choose a window size divisible by the denominator of the interleaving ratios because our tests show it improves convergence.

\noindent\textbf{Evaluation Metrics.} 
Our main performance metrics are privacy—measured as robustness to \acrshort{dlg} attacks—and model accuracy, defined as the percentage of correctly classified test samples after convergence. We also track execution time, including client training, server aggregation, and encryption/decryption, to evaluate computational cost. Additionally, we measure the total transmitted bytes until convergence to assess bandwidth consumption.

To evaluate privacy, we assume the attacker aims to leak training data using the \acrshort{dlg} attack~\cite{DLG} to recover authentic images.
To quantify to which extent the attacker is successful, we use Universal Quality Index~(UQI) \cite{DBLP:journals/spl/WangB02}, Multiscale Structural Similarity (MSSSIM)~\cite{MSSSIM}, and Visual Information Fidelity (VIF)~\cite{DBLP:journals/tip/SheikhB06} to measure the similarity between authentic and recovered images, in accordance with the evaluation criteria in~\cite{DBLP:journals/corr/abs-2303-10837}.
We consider that 100 authentic images of each class are attacked 
and calculate image similarity scores based on the above metrics.
We document each metric's highest score achieved during any \acrshort{dlg} attack.
Additionally, to see if authentic images can be recovered from a synthetic model, \eg the models from synthetic rounds, we perform the \acrshort{dlg} attack on synthetic images in a similar manner. 
However, in this case, we compare each recovered image against all authentic images.
We record the similarity scores for the images with the highest MSSSIM scores since MSSSIM appears to best reflect perceived image similarity.
We then document the average similarity scores from the recorded scores.

\noindent\textbf{Benchmarks.} We compare \acrshort{nameNovel} to the following baselines:

\noindent $\bullet$  \acrshort{fe}, \ie  no interleaving is used and all models are sent fully encrypted using \acrshort{he} ($\eta{=}1$);

\noindent $\bullet$ \acrshort{s-he}, \ie no interleaving is used and all models  are sent encrypted using Selective \acrshort{he} with $\eta{=}0.2$.

\section{Results and Discussion}
\label{sec:results}

\begin{figure}[t!]
    \centering
    \subfigure[Auth.]{\includegraphics[width=0.15\linewidth]{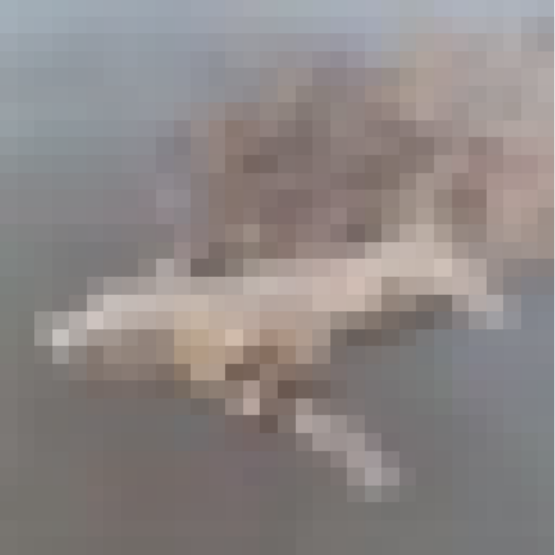}}
    \hspace{2mm}
    \subfigure[$\eta{=}0$]{\includegraphics[width=0.15\linewidth]{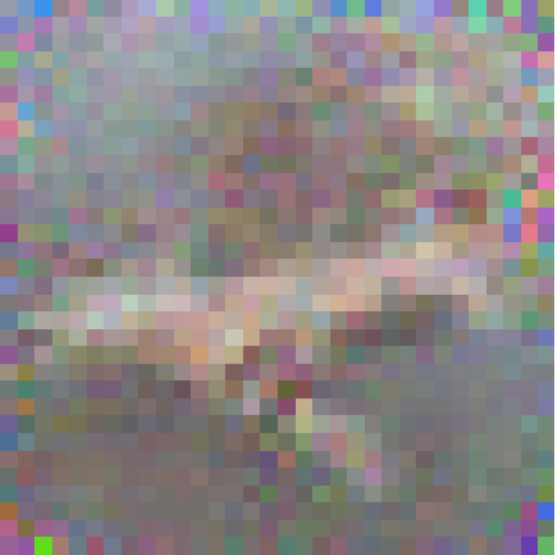}}
    \hspace{2mm}\subfigure[$\eta{=}0.1$]{\includegraphics[width=0.15\linewidth]{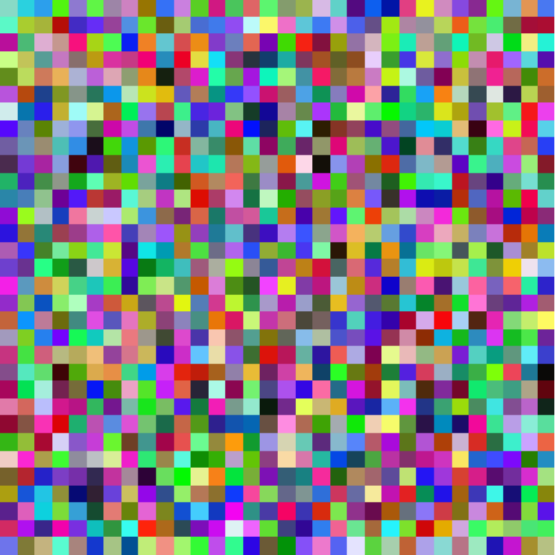}}
    \hspace{2mm}\subfigure[$\eta{=}0.2$]{\includegraphics[width=0.15\linewidth]{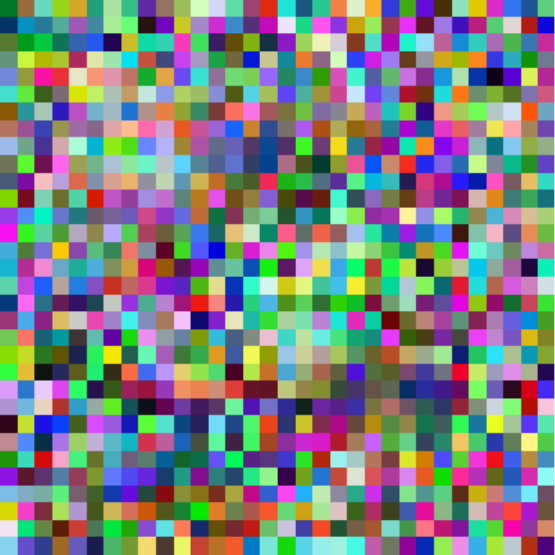}}
    \caption{Result of the \acrshort{dlg} attack on a specific authentic image (a) when models are sent  unencrypted (b), and  encrypted with $\eta{=}0.1$ (c) and $0.2$ (d).}
    \label{fig:img_dlg_examples}
\end{figure}

\noindent\textbf{Privacy Protection.}
To gain insight into the significance of the similarity metrics, we first conduct experiments along the lines of the image classifier methodology in~\cite{DBLP:journals/corr/abs-2303-10837}. 
By varying the encryption ratio $\eta$ between 0 (no encryption) and 1 (fully encrypted models), we achieve similar scores to those in \cite{DBLP:journals/corr/abs-2303-10837} in terms of image similarity when protecting against \acrshort{dlg} attacks. 
Result samples  in \Cref{tab:dlg_results}  present the maximum similarity scores obtained during the \acrshort{dlg} attacks. 
Consistently with~\cite[Fig.~9] {DBLP:journals/corr/abs-2303-10837}, 
our results reveal that, for $\eta{=}0.1$, all encrypted cases score substantially lower (hence, better) than the  unencrypted, further demonstrating
that our embedding of Selective \acrshort{he} into \acrshort{fedavg} protects against the attack.  
Importantly, these values also provide a baseline against which we can compare the similarity scores obtained under \acrshort{nameNovel}. 
To illustrate the effects of Selective \gls{he}, \Cref{fig:img_dlg_examples} presents the recovered images from \acrshort{dlg} attacks on a specific authentic image, when the models are not encrypted, as well as when they are encrypted with $\eta{=}0.1$ and 0.2. 
While an unencrypted model leaks the training image, Selective \gls{he} can effectively prevent privacy leakage despite a relatively low encryption~ratio.

\begin{table}[t!]
\centering
\caption{Similarity metrics for \acrshort{dlg} attack when data is not  encrypted ($\eta{=}0$), and when Selective \acrshort{he} is applied with different encryption ratios (no interleaving is used, \ie $\rho{=}0$)}
\label{tab:dlg_results}

\begin{tabular}{@{}llllllllllll@{}}
\toprule
       & \multicolumn{1}{c}{Unencrypted} & \multicolumn{1}{c}{$\eta{=}0.1$} & \multicolumn{1}{c}{$\eta{=}0.2$}  \\ \midrule
UQI & 0.9954 & 0.8950 & 0.8712  \\
MSSSIM & 0.9640 & 0.7000 &  0.6407 \\
VIF & 0.3167 & 0.0629 &  0.0602 \\ \bottomrule
\end{tabular}
\end{table}

\begin{table*}[t!]
\caption{\acrshort{nameNovel} ($\eta{=}0.2$): Avg.\,similarity scores during \acrshort{dlg} attacks for different synthetic rounds (between brackets) and values of the interleaving ratio. Values between brackets report the increase in percentage 
w.r.t. \acrshort{s-he}, where green text indicates better privacy and red indicates worse privacy}
\label{tab:DLG_ALt-FL}
\centering
\begin{tabular}{@{}lcccccc@{}}
\toprule
& \multicolumn{1}{c}{$\rho{=}0.5$ (40)} & \multicolumn{1}{c}{$\rho{=}0.5$ (80)} & \multicolumn{1}{c}{$\rho{=}0.5$ (100)} & \multicolumn{1}{c}{$\rho{=}0.25$ (40)} & \multicolumn{1}{c}{$\rho{=}0.25$ (80)} & \multicolumn{1}{c}{$\rho{=}0.25$ (100)} \\ \midrule
UQI & 0.782 ($\color{OliveGreen}-10.3\%$) & 0.779 ($\color{OliveGreen}-10.6\%$) & 0.780 ($\color{OliveGreen}-10.5\%$) & 0.781 ($\color{OliveGreen}-10.3\%$) & 0.782 ($\color{OliveGreen}-10.2\%$) & 0.779 ($\color{OliveGreen}-10.6\%$) \\
MSSSIM & 0.537 ($\color{OliveGreen}-16.1\%$) & 0.524 ($\color{OliveGreen}-18.2\%$) & 0.528 ($\color{OliveGreen}-17.6\%$) & 0.539 ($\color{OliveGreen}-15.9\%$) & 0.527 ($\color{OliveGreen}-17.7\%$) & 0.528 ($\color{OliveGreen}-17.6\%$) \\
VIF & 0.044 ($\color{OliveGreen}-26.8\%$) & 0.042 ($\color{OliveGreen}-29.7\%$) & 0.042 ($\color{OliveGreen}-29.8\%$) & 0.044 ($\color{OliveGreen}-27.2\%$) & 0.043 ($\color{OliveGreen}-29.2\%$) & 0.042 ($\color{OliveGreen}-29.6\%$)\\ \bottomrule
\end{tabular}
\end{table*}

\begin{figure}[t!]
    \centering
    \subfigure[]{\includegraphics[width=0.15\linewidth]{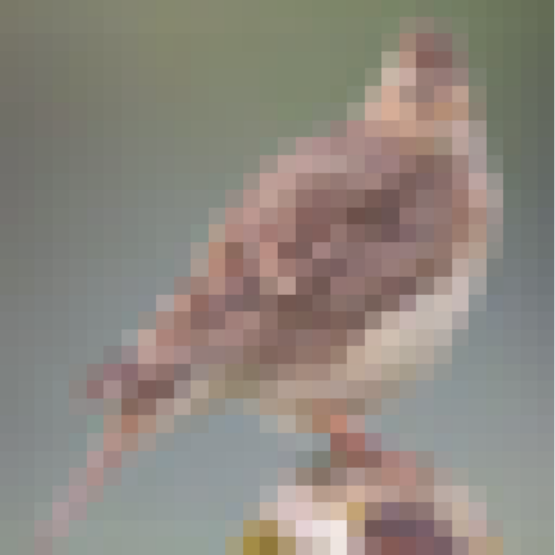}}
    \subfigure[]{\includegraphics[width=0.15\linewidth]{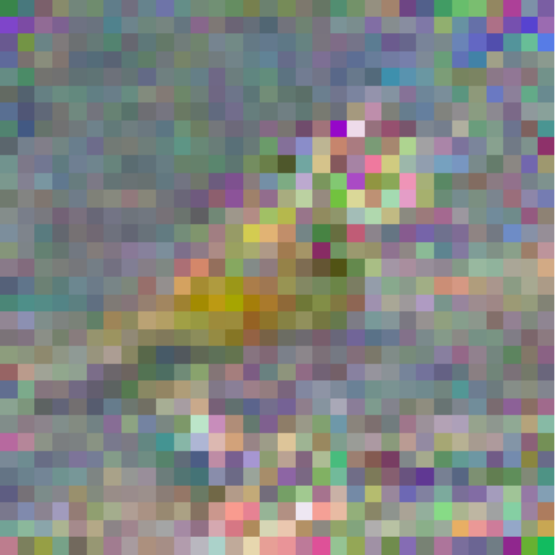}}
       \subfigure[]{\includegraphics[width=0.15\linewidth]{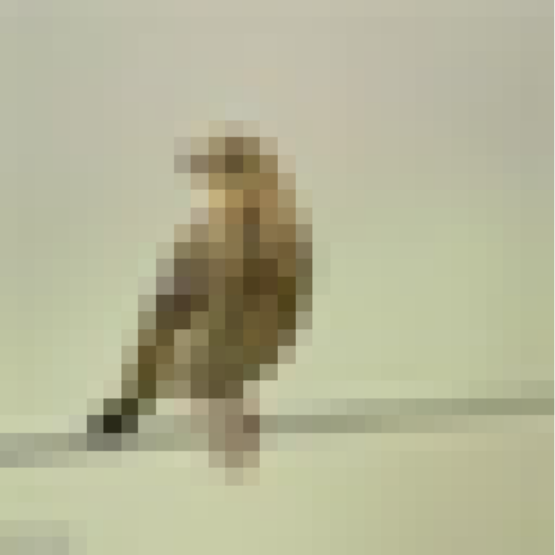}} \subfigure[]{\includegraphics[width=0.15\linewidth]{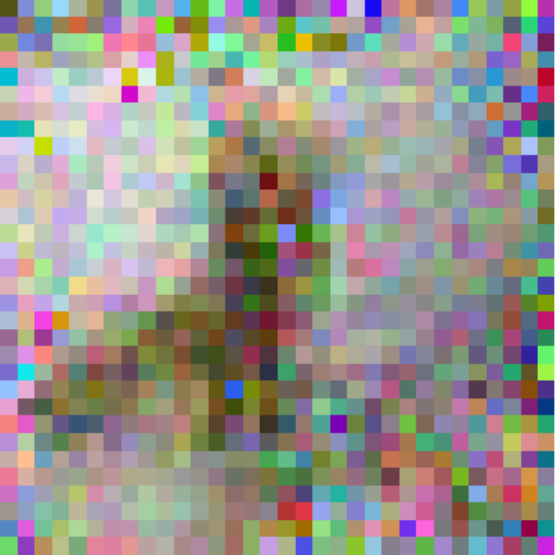}}
          \subfigure[]{\includegraphics[width=0.15\linewidth]{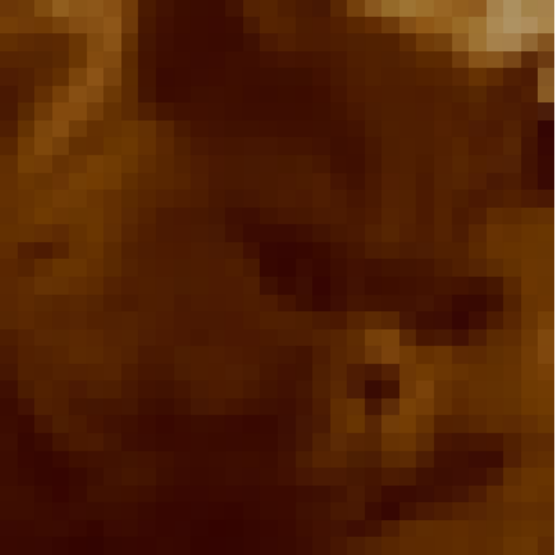}}
    \subfigure[]{\includegraphics[width=0.15\linewidth]{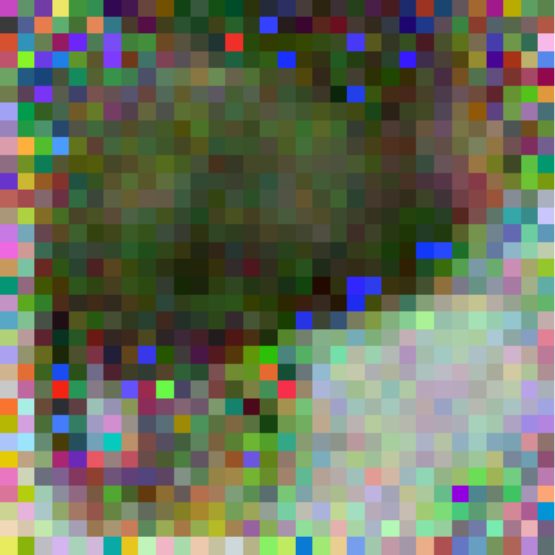}}  
    \caption{Attack launched on synthetic  rounds when $\rho{=}0.5$: exemplary   authentic images (a), (c), (e) and corresponding recovered image for (resp.) max UQI (b), max MSSSIM (d), and max VIF (f).}
    \label{fig:img_syn_dlg_examples}
      \BBB
\end{figure}

Focusing now on \acrshort{nameNovel}'s performance, \Cref{tab:DLG_ALt-FL} presents the average similarity scores the proposed solution yields when \acrshort{dlg} attacks are launched on different synthetic rounds, for  $\rho{=}0.25, 0.5$ (note that the higher the similarity score, the more successful the attack).
In all synthetic rounds shown in \Cref{tab:DLG_ALt-FL}, the average similarity scores using \acrshort{nameNovel} are lower than those using Selective \acrshort{he} in \Cref{tab:dlg_results}. This suggests that synthetic rounds perform comparably to \acrshort{s-he}.  
Our closer inspection of the data reveals that the average scores and those of the most successful attacks are not significantly different.
This good performance can also be observed by looking at \Cref{fig:img_syn_dlg_examples}: even for the highest similarity scores, the recovered images differ greatly from the authentic ones.

\begin{table*}[t!]
\centering
\BB
\caption{Averaged training costs for \acrshort{nameNovel} and its benchmarks: no.\,of rounds needed for model convergence, total no.\,of ciphertext Mbytes, per-client  total no.\,of transmitted Mbytes and computational time, and achieved accuracy level. Values between brackets report the increase in percentages w.r.t. \acrshort{s-he}, where the green text indicates better performance and red indicates worse performance}
\label{tab:total_costs}
\begin{tabular}{@{}rccccc@{}}
\toprule
            &
            \multicolumn{1}{c}{Convergence\, [\#rounds]} &
            \multicolumn{1}{c}{Ciphertext\,[MB]} & 
            \multicolumn{1}{c}{Client tot. comm.\,cost\,[MB]} & 
            \multicolumn{1}{c}{Client tot. comp. time\,[s]} & \multicolumn{1}{c}{Accuracy\,[\%]} \\ \midrule
\emph{S-HE}, $\rho{=}0$ & 77 &  0.14 & 2.25 & 3340 & 55.58 \\
\acrshort{nameNovel}, $\rho=0.25$ & 91 ($\color{BrickRed}+18.4\%$) & 0.13 ($\color{OliveGreen}-8.3\%$)   &  2.62 ($\color{BrickRed}+16.7\%$)  & 3967 ($\color{BrickRed}+18.8\%$)  & 61.04 ($\color{OliveGreen}+9.8\%$) \\
\acrshort{nameNovel}, $\rho=0.5$ & 92 ($\color{BrickRed}+20.5\%$) & 0.08 ($\color{OliveGreen}-39.1\%$) & 2.62 ($\color{BrickRed}+16.8\%$) & 4036 ($\color{BrickRed}+20.8\%$) & 63.02 ($\color{OliveGreen}+13.4\%$)  \\ \bottomrule
\end{tabular}
   \BBB
\end{table*}

\begin{figure}[ht]
    \centering
    \includegraphics[width=0.8\linewidth]{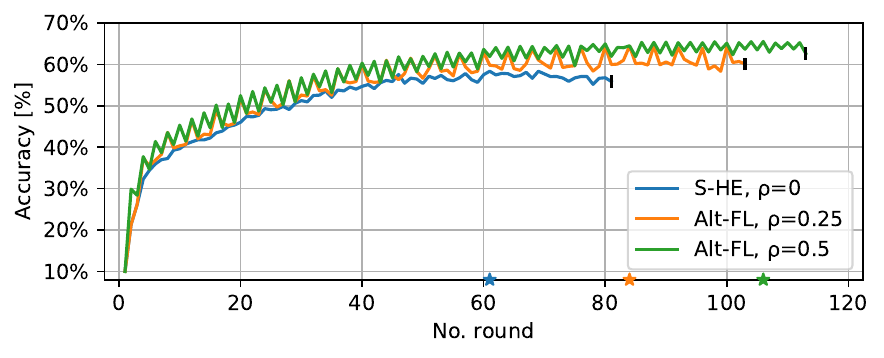}
    \BBB
    \caption{The LeNet5 model's accuracy of \acrshort{nameNovel} and its benchmarks until convergence. Stars markers denote the round at which schemes reached peak accuracy.}
    \label{fig:acc_enc_0_2}
    \BB
\end{figure}

\noindent\textbf{Accuracy and Overhead.}
\Cref{fig:acc_enc_0_2} shows the accuracy  achieved by \acrshort{nameNovel} with $\rho{=}0.25, 0.5$ and Selective \acrshort{he} with no interleaving (S-HE, $\rho{=}0$) and $\eta{=}0.2$, for a single experiment execution. 
Notice that the scheme FE, $\rho{=}0$  yields the same accuracy level   as S-HE, $\rho{=}0$, and, thus,  is omitted for brevity. 
We also omit models with higher encryption ratios since they achieve the same accuracy level. 
Importantly, one can observe that \acrshort{nameNovel} substantially increases accuracy compared to when only authentic rounds are performed (up to 13.4\% w.r.t. S-HE, $\rho{=}0$ using the same authentic dataset).

\begin{figure}[ht]
    \BB\centering
    \includegraphics[width=0.8\linewidth]{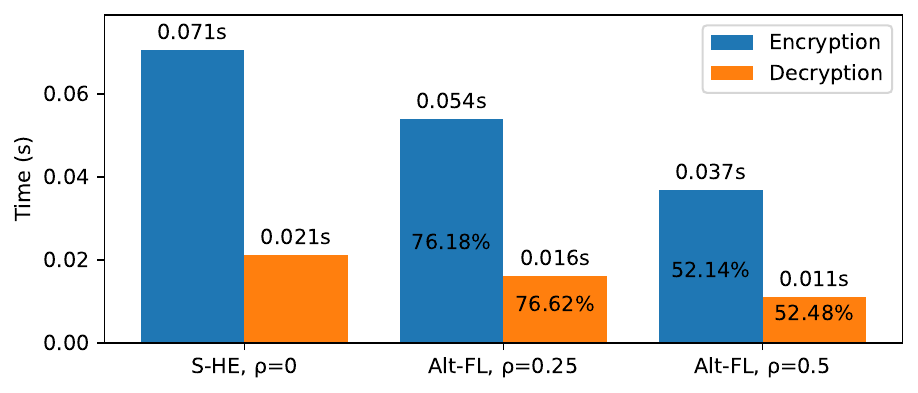}
    \BBB
    \caption{Encryption and decryption time with encryption ratio~$\eta{=}0.2$.}
    \label{fig:enc_dec_times}
    \BB
\end{figure}

Next, we focus on the computational and communication costs of encryption.
\Cref{fig:enc_dec_times} illustrates the average encryption and decryption time for \acrshort{nameNovel} for different values of $\rho$, compared to S-HE, $\rho{=}0$  (the percentage of the time required by \acrshort{nameNovel} w.r.t. S-HE, $\rho{=}0$ is highlighted within each bar). 
It is evident that an increase in the fraction of synthetic rounds leads to a reduction in the encryption/decryption time introduced by \acrshort{he}, which is roughly equivalent to the value of $\rho$, leading to a 48\% decrease in overhead.

Finally, \Cref{tab:total_costs} shows \acrshort{nameNovel}'s communication and computation training costs with $\rho{=}0.25, 0.5$, compared to the benchmarks when $\eta{=}0.2$.  
As observed above,
\acrshort{nameNovel} requires more time for the model to converge because synthetic rounds use data different from the authentic dataset.
Notably, on average, the model converges after 77 rounds under S-HE, $\rho{=}0$, but it takes 91 and 92 rounds with $\rho{=}0.25$ and $\rho{=}0.5$, resp. 
Despite the longer convergence time, the total ciphertext transmitted during training decreases by 8.3\% and 39.1\% for $\rho{=}0.25$ and $\rho{=}0.5$, resp., compared to the benchmarks. 
Thus, although \acrshort{nameNovel} increases convergence time, it can effectively reduce HE-related costs and improve accuracy.

\section{Conclusion}  

We enhanced \gls{fedavghe} by introducing \acrshort{nameNovel} to address relevant challenges in \acrshort{fl}: increasing model accuracy and safeguarding privacy while mitigating the overhead associated with \acrshort{he}. These enhancements leverage an interleaving strategy that alternates between authentic and synthetic training rounds, which are transmitted in plaintext.
Our experiments show that \acrshort{nameNovel} exhibits a moderate increase in computation time compared to the benchmark \acrshort{s-he}. Such an increase is thus worth it, as \acrshort{nameNovel} delivers significant accuracy improvements (by 13.4\%) and a substantial decrease in HE-related computational costs (by up to 48\%), while preserving privacy.

\bibliographystyle{IEEEtran}
\bibliography{references_short}

\end{document}